# Segmentation of diagnostic tissue compartments on whole slide images with renal thrombotic microangiopathies (TMAs)

Huy Q. Vo,[1] Pietro A. Cicalese,[1, 2] Surya Seshan,[3] Syed A. Rizvi,[1] Aneesh Vathul,[1] Gloria Bueno,[4] Anibal Pedraza Dorado,[4] Niels Grabe,[5] Katharina Stolle,[6] Francesco Pesce,[7] Joris J.T.H. Roelofs,[8] Jesper Kers,[8] Vitoantonio Bevilacqua,[9] Nicola Altini,[9] Bernd Schröppel,[10] Dario Roccatello,[11] Antonella Barreca,[12] Savino Sciascia,[11] Chandra Mohan,[2]* Hien V. Nguyen,[1]* Jan U. Becker[13]*

[1] Department of Biomedical Engineering, University of Houston, Houston, USA
[2] Biomedical Engineering & Medicine, University of Houston, Houston, TX, USA.
[3] Department of Pathology, Weill-Cornell Medical Center/New York Presbyterian Hospital, New York, NY, USA
[4] VISILAB Research Group, University of Castilla–La Mancha, Ciudad Real, Spain
[5] Steinbeis Transfer Center for Medical Systems Biology, Heidelberg, Germany
[6] Department of Internal Medicine, Faculty of Medicine, University Bonn, Bonn, Germany
[7] Division of Renal Medicine "Fatebenefratelli Isola Tiberina – Gemelli Isola", Rome, Italy
[8] Department of Pathology, Amsterdam UMC, University of Amsterdam, Amsterdam, the Netherlands
[9] Department of Electrical and Information Engineering (DEI), Polytechnic University of Bari, 70126 Bari, BA, Italy
[10] Section of Nephrology, University Hospital Ulm, Ulm, Germany
[11] Coordinating Center of the Interregional Network for Rare Diseases of Piedmont and Aosta Valley, San Giovanni Bosco Hub Hospital, Turin, Italy, University Center of Excellence on Nephrologic, Rheumatologic and Rare Diseases with Nephrology and Dialysis Unit and Center of Immuno-Rheumatology and Rare Diseases (CMID),
Sektion Nephrologie, Klinik für Innere Medizin I, Universität Ulm, Ulm, Germany
[12] Division of Pathology, Città della Salute e della Scienza Hospital
[13] Institute of Pathology, University Hospital of Cologne, Cologne, Germany

* Authors contributed equally

## Abstract

The thrombotic microangiopathies (TMAs) manifest in renal biopsy histology with a broad spectrum of acute and chronic findings. Precise diagnostic criteria for a renal biopsy diagnosis of TMA are missing.
As a first step towards a machine learning- and computer vision-based analysis of wholes slide images from renal biopsies, we trained a segmentation model for the decisive diagnostic kidney tissue compartments artery, arteriole, glomerulus on a set of whole slide images from renal biopsies with TMAs and Mimickers (distinct diseases with a similar nephropathological appearance as TMA like severe benign nephrosclerosis, various vasculitides, Bevacizumab-plug glomerulopathy, arteriolar light chain deposition disease). Our segmentation model combines a U-Net-based tissue detection with a Shifted windows-transformer architecture to reach excellent segmentation results for even the most severely altered glomeruli, arterioles and arteries, even on unseen staining domains from a different nephropathology lab. With accurate automatic segmentation of the decisive renal biopsy compartments in human renal vasculopathies, we have laid the foundation for large-scale compartment-specific machine learning and computer vision analysis of renal biopsy repositories with TMAs.

## Index Terms

Machine learning, nephropathology, histology, TMA, thrombotic microangiopathy, artery, arteriole

## Introduction

The thrombotic microangiopathies (TMAs) are a heterogenous class of diseases, often with renal involvement. The umbrella term TMA coined by Symmers [1] contains thrombotic-thrombocytopenic purpura (TTP) and hemolytic uremic syndrome (HUS) [2]). In contrast to TTP, caused by defects in or autoantibodies against ADAMTS13 [3, 4], HUS affects the kidney rather than the central nervous system. HUS can present as typical HUS which is caused by shiga toxin and rarely biopsied. Atypical HUS (aHUS) is mainly attributed to defects in complement regulation, either acquired or genetic. However, other factors can also trigger or cause aHUS, particularly medication and infections [5, 6].

For the clinical diagnosis of aHUS, two largely similar algorithms have been proposed. Neither of these two algorithms incorporate nephropathological findings. [7, 8]. In fact, nephropathology has contributed little to the field of aHUS or TMA in the recent decades. This can be attributed to the lack of comprehensive clinicopathological datasets, correlating histopathology with etiological data and outcome, which has markedly improved with modern complement inhibition.

In nephropathology, the last 5 years have seen the introduction of machine learning to the analysis of what are now virtual, scanned glass slides or whole slide images (WSIs) [9, 10]. This computer vision approach is very promising for the integration of nephropathology in precision medicine for several reasons: it offers perfect reproducibility and it is the only realistic option to analyse thousands of biopsies in a hypothesis-free approach. Machine learning could integrate histopathology, the clinical tabulations at the time of biopsy and even genetic code to determine the three most important outputs: diagnosis, prognosis and treatment response [10].

Individual WSIs (each level section about 250 MB at x 40 resolution with up to 6 on a single slide at x40 resolution), are too large an input for ML models. Although multiple-instance learning architectures can use randomly tiled WSIs as the output of diagnostic classifiers [11, 12], we prefer to follow nephropathology concepts in our approach. A central nephropathological concept is the division of renal tissue into the compartments artery, arteriole, glomerulus, arteriole, cortex tubulointerstitium, medulla tubulointerstitium. In the TMAs, the first three former compartments artery, arteriole and glomerulus are directly affected and thus the most important. In this manuscript we describe the development of an instance segmentation model for these three compartments capable of accurate predictions on all four main nephropathological stainings hematoxylin-eosin, periodic acid-Schiff (PAS), Jones silver, trichrome even with severe alterations by vascular disease.

## Methods

**Patients and Biopsies**

All n=60 biopsies were diagnosed by experienced nephropathologists in Cologne (JUB), at Weill-Cornell (SuS) and in Turin (AB). The TMA cohort consisted of n=29 biopsies with a histopathological diagnosis of TMAs (n=21 from Cologne, n=5 from Turin and n=3 from Weill-Cornell). Inclusion criteria for TMAs included microthrombi, split glomerular basement membranes without any other apparent cause, dysoria of arteries and arterioles, with intra- or extramural leakage of blood constituents, foam cell, myxoid and hypoelastotic intimal changes as well as onion-skin transformation and obliteration of arteries and arterioles. The aetiology of TMA included arterial hypertension, bacterial infection, systemic sclerosis, anti-phospholipid antibody disease, chemotherapy/Avastin treatment, and aHUS of unknown aetiology, The second cohort of n=31 Mimickers consisted of severe renal vasculopathies (n=25 from Cologne, n=5 from Turin, n=1 from Weill-Cornell). Mimickers were biopsies with a diagnosis showing similar histological features to TMA, often resulting in severe distortions of the compartment morphology. These diseases included severe benign nephrosclerosis (n=16), anti-neutrophil cytoplasmic antibody (ANCA) associated glomerulonephritis with leukocytoclastic arteritis (n=10), cryoglobulinemic glomerulonephritis/arteritis (n=3), Bevacizumab-associated obliterative glomerulopathy [13] (n=1), light chain deposition disease with arteriolopathy (n=1). The training set for the segmentation model consisted of n=40 biopsies from the archives of the Institute of Pathology, University Hospital of Cologne. The validation set consisted of n=10 biopsies, n=4 biopsies from the Department of Pathology, Weill-Cornell Medical Center plus n=6 from Cologne. The test set for the segmentation model consisted of n=10 biopsies, n=5 with TMA and n=5 Mimickers from the University Hospital of Turin. We included hematoxylin-eosin (HE), periodic-acid Schiff (PAS), Jones methenamine silver, and trichrome stainings (including Masson's, Elastica-van Gieson and acid-fuchsin orange G) in this project.
The biopsies from Cologne and Turin were scanned with a Hamamatsu Scanner (Hamamatsu Photonics, Herrsching am Ammersee, Germany); the biopsies from Weill-Cornell were scanned with an Aperio GT 450 scanner (Leica Biosystems, Deer Park, IL, USA). All scans were obtained with a x40 objective.

**Expert Annotation (Segmentation)**

All expert annotations (manual segmentations) were performed under QuPath [14] on a Wacom Cintiq Pro 16-inch pen display (Wacom, Düsseldorf, Germany) by an experienced nephropathologist (JUB). Object classes included Glomerulus, Arteriole, Artery, Cortex, Medulla, Capsule/Other, Ignore. For arterial and arteriolar annotations, we included the walls up to the outer border of the media as well as the lumina; all luminal areas in incompletely captured, "opened" segments. Arteries were distinguished from Arterioles by the number of smooth muscle cell layers, with at least 2 defining an Artery [15]. Glomeruli were annotated including the tuft, Bowman's space and the perimeter of Bowman's capsule. Empty Bowman's capsules were disregarded for glomerular annotation. Dislodged glomerular tufts were included; tufts, if still attached to the capsule, were included with the capsule segment; if completely detached, under inclusion of the entire tuft only. Ignore was used for artefacts such as tissue, folds or dirt. In total, n=2,439, n=15,399 and n=10,892 objects were annotated for the three classes Artery, Arteriole, Glomerulus. Cortex was differentiated from Medulla by the characteristic tubular morphology and by the presence of glomeruli in the former and the characteristic parallel alignment of tubules/ducts and the absence of arteries in the latter; Capsule/Other contained renal capsule and other non-renal tissue elements such as connective tissue, skeletal muscle and other incidental tissue/organ elements contained in the biopsy.

**Segmentation Module Development**

Our segmentation module is comprised of 3 main stages (tissue segmentation, instance segmentation and post-processing), the entire workflow is shown in Figure 1.

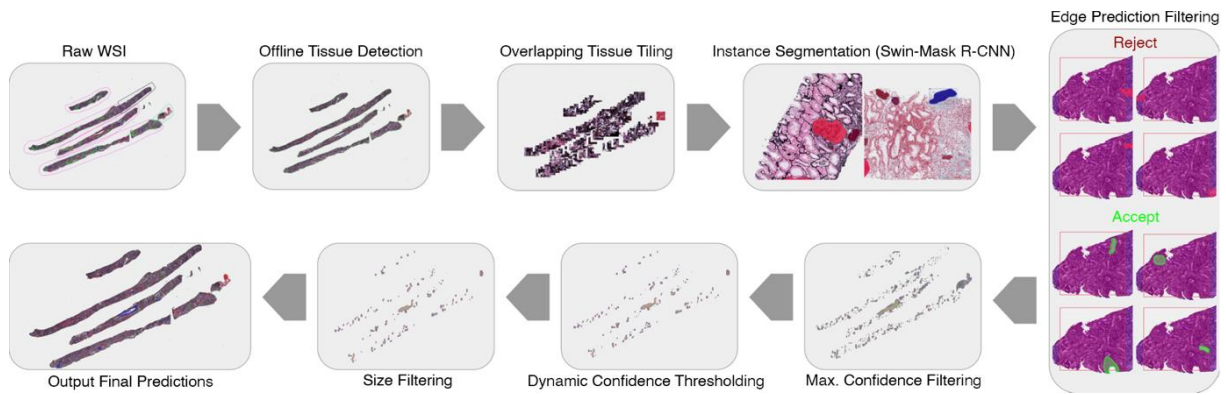

**Figure 1:** Schematic overview of the segmentation model. We first performed tissue detection on our WSI, followed by strided, overlapping tiling to allow us to account for edge artefacts. We then passed each tile through our Swin Transformer backbone Mask R-CNN instance segmentation architecture; for each tile, we used edge-prediction filtering for all predictions. Then, we took the maximum instance pixel confidence value per class, and then performed dynamic thresholding to eliminate noise. Finally, we filtered instances with a binary mask area of less than 25 pixels, and mapped our final cleaned predictions to the WSIs.

*1/ Tissue segmentation:*
Figure 1 shows a schematic overview of the full segmentation module. We first performed tissue segmentation on our WSIs. Tissue segmentation is a binary segmentation problem where a computer vision technique tries to discriminate between tissue region and background region. Particularly with the background silver impregnation of glass slides in Jones stainings, offline tissue detection (or segmentation) by thresholding by methods such as Otsu's [16] did not seem sufficient. Hence, we resorted to a deep learning model for this initial step of tissue segmentation. Because different tissue regions (or background regions) are not necessarily discriminated into different instances, we choose semantic segmentation for tissue segmentation. To this end, we trained a lightweight variant of the U-Net architecture [17] to segment tissue from WSI background pixels using aforementioned tissue annotations (Capsule/Other, Cortex, Medulla). We leverage the available annotations of these 3 tissue types, because they can cover most of the tissue regions in a WSI, which consequently helps the model to learn to segment tissue regions from background regions. As a result, the number of segmentation classes is 4: Capsule/Other, Cortex, Medulla, and background. At the inference stage, the segmented regions of 3 different tissue types will be aggregated into one tissue mask. The resulting binary tissue mask will have white colour for tissue regions and black colour for background regions.

*2/ Instance segmentation of glomerulus, arteriole, artery:*
   a) Training deep instance segmentation model:

Instance segmentation is used not only to segment tissues of interest but also to discriminate different instances of the same tissue class (glomerulus, artery, arteriole) in a whole slide image. To train the model, we use tiles as inputs, not crops. The difference between tile and crop is that a tile is extracted from a grid overlayed on a WSI, while crop closely surrounds a tissue of interest (glomerulus, arteriole, artery). The size of tile is usually bigger than crop and thus each tile can contain more than one tissue instance. We choose to use tile because this will fit with the practical scenario (the inference stage) where the model will receive tiles from a WSI as input. We then passed each tile through our adaptation of Mask R-CNN [18] instance segmentation architecture using a shifted windows (Swin) Transformer [19] as backbone. Each predicted tissue instance can then be utilised to serve as one training sample (one crop) for subsequent deep learning tasks such as image classification or retrieval.

   b) Post-processing steps in the inference (prediction) stage:

Predicted tissue crops of glomeruli (or arterioles, arterys), retrieved from the trained model of step (a) in the inference stage, contain a great amount of "noise" crops. Thus, we propose a multi-step post-processing stage for dealing with these "noise" crops. All of the challenges ("noise") and the corresponding solutions are described below. For each tile, we excluded predictions fulfilling both of the following two criteria: 1) more than 20% of the mask circumference were on the tile edge, and 2) more than 90% of the mask area were within the outer 10% of the tile frame. After this edge prediction filtering, we took the maximum instance pixel confidence value per class. For overlapping predictions belonging to the same class, we simply selected the highest confidence instances from our raw instance-level pixel confidence scores given their superior quality. Higher confidence predictions tended to have more surrounding context in their respective tile when compared to adjacent tiles, resulting in better segmentation masks. Visual inspection revealed that our WSI set carried significant domain shifts not only due to the four different stainings HE, Jones, trichrome and PAS, but also for individual staining between the three institutions. We speculated that these problematic domain shifts could be overcome with dynamic confidence thresholding (DCT), eliminating noise. The rationale behind DCT was as follows: typically, instance segmentation predictions are filtered using single confidence thresholds unique to each class. Dynamic adaptation of this thresholding might deliver better results on the diverse domains of our WSI set. For DCT, we trained a multilayer perceptron (MLP) regression model to map the following to an ideal confidence threshold (using maximum F1 confidence thresholds as ground truths): 1) Binned unique predicted confidence values, 2) binned predicted confidence value frequencies and 3) one-hot encoded class vectors. We trained our DCT network on WSIs outside of the current evaluation set (e.g. training and validation WSIs when predicting confidence thresholds for testing). Finally, we filtered instances with a binary mask area of less than 25 pixels, and then mapped our cleaned predictions for the three instance classes Glomerulus, Arteriole, Artery to the WSI. Although infrequent, we opted to remove overlapping prediction masks belonging to different classes (as opposed to selecting the higher confidence class) by using a mask IoU cut-off of 0.7 computed between the overlapping prediction masks.

**Datasets**

For tissue segmentation, in total there are 666 WSIs (samples), in which 538 samples are used for training and 128 samples are used for validation. To be able to do the training, both the WSIs and their corresponding tissue masks were resized to thumbnails of resolution 4096 x 4096. For tissue type segmentation of glomerulus, arteriole, artery, the next step of instance segmentation used 32,732 extracted tiles with a size of 4096 x 4096 pixels at 20x resolution, using a QuPath Groovy script (https://gist.github.com/scottdoy/62b7413db1993c9bb6513f4ff7d3860f; last accessed November 2023), with a stride of 32 pixels to avoid edge artefacts. All tiles are then resized to 2048 x 2048 to match the required input image size of the Swin Transformer. About data splitting, 26,924 tiles are used for training (set 1), and 5808 tiles are used to choose the most optimal model (set 2). The data splitting is carried out in a way so that case ids (patient ids) in set 1 and set 2 are totally different. In details, there are 40 case ids in the the first set and 10 case ids in the second set. In set 1, there are 20 case ids assigned with TMA and 20 case ids assigned with Mimicker. In set 2, there are 5 case ids assigned with TMA and 5 case ids assigned with Mimicker. Note that the labels of TMA or Mimicker are not used in all our segmentation experiments. They are solely employed to ensure a fair data split, which, in turn, contributes to reliable experimental results. For further clarification, it is worth mentioning that each case ID may be associated with more than one slide ID. Consequently, the training set comprises 346 slide IDs, while the set for choosing the most optimal model includes 57 slide IDs.

Moreover, because each tile may have more than one annotation (one tissue instance), a tile may contain all three tissue types, and for each tissue type there can be more than one instance. In 26,924 tiles of the training set, there are a total of 16,2567 annotations: 72,008 glomerulus instance annotations, 71,104 arteriole instance annotations, 19,455 artery instance annotations. In 5808 tiles for choosing the most optimal model, there are a total of 41,014 annotations: 1,9535 glomerulus instance annotations, 17,696 arteriole instance annotations, 3783 artery annotations. Note that the numbers listed above are not the true original number of glomeruli, arterioles, and arteries annotated in all WSIs. Because tiles are extracted from grid, then there are cases that a tile only covers a partial glomerulus (or arteriole, artery) and the other neighbour tiles will cover the remains of that glomerulus (or arteriole, artery). These different parts of a glomerulus are considered as different instances. And thus, an original annotation of a tissue may become multiple instance annotations. This helps explain why there will be more instance annotations than the true number of original annotations. The numbers of original annotations are listed below. The training set for these three classes consisted of n=7694 Glomerulus crops, n=10836 Arteriole and n=1658 Artery crops (all from Cologne). The validation set consisted of the n=2200 Glomerulus, n=3099 Arteriole and n=532 Artery crops, all from the Weill-Cornell biopsies. The test set consisted of n=998 Glomerulus, n=1464 Arteriole and n=249 Artery crops, all from Turin

**Training settings**
*1/ Details about training the tissue semantic segmentation model:*
We use the AdamW optimizer [20] with the initial learning rate of 1e-4, and the learning rate is reduced if the evaluation Dice score does not show any improvement every 100 epochs. We train the model in 500 epochs with batch size equal 6 using the cross-entropy loss. In each training iteration, 4 data augmentation techniques are used to increase the number of training samples: horizontal flipping, vertical flipping, channel shuffling, random cropping with crop width randomly selected from quarter to half of the original size (and the same for height). All the augmentation types are set to operate randomly at ratio 0.5.

*2/ Details about training the instance segmentation model for 3 classes of glomerulus, arteriole, and artery:*
We use the mmdetection library (https://github.com/open-mmlab/mmdetection ; last accessed November 2023) in our experiments. The tiny version of Swin Transformer is used as backbone for the Mask R-CNN architecture. The whole model is trained with the maximum number of epochs set as 20 and the initial learning rate is 0.0001. AdamW is also used as optimizer with weight decay equal to 0.1. We use step learning rate scheduler with the learning rate decreased by a factor of 10 at epoch 13 and epoch 16. Linear warmup is employed for the first 1000 training iterations with the warmup ratio is set to 0.001. The cross-entropy loss is utilized in both the localization and segmentation components. The input image resolution is 2048x2048. Training batch size is 3 per GPU for 4 Quadro RTX 8000. For data augmentation, AutoAugment [21] is used with resize policy having 11 different image scales: (768x2048), (896x2048), (1024x2048), (1152x2048), (1280x2048), (1408x2048), (1536x2048), (1664x2048), (1792x2048), (1920x2048), (2048x2048). We also use random flip with ratio 0.5.

**Evaluation metrics**
Performance metrics for the instance segmentation included intersection-over-union (IOU also known as Jaccard index), average precision (AP, also known as positive predictive value), average recall (AR, also known as sensitivity), F1 score (F1, also known as the harmonic mean of precision and recall) and average specificity (AS).

## Distribution of Code

The code is available here: https://github.com/hula-ai/kidney-wsi-seg.

## Ethical Permissions

The retrospective use of de-identified data for epidemiological research projects like this in Cologne is permitted by state law in North Rhine-Westphalia (Berufsordnung der nordrheinischen Ärztinnen und Ärzte), Germany (https://www.aekno.de/fileadmin/user_upload/aekno/downloads/2022/berufsordnung-2021.pdf; last accessed November 2023).

# Results

### Segmentation Module Performance

We show that our system produces strong performance for all three object classes, artery, arteriole and glomerulus, TMA and Mimicker, for all four main paraffin nephropathology stainings. The strong performance persisted even on previously unseen staining domains (see Table 1 and 2).

**Table 1.** Dynamic thresholding performance when compared to static thresholds for each class. We noted that dynamic thresholding performed similarly or better than static thresholding without using ground truth knowledge. Also listed is the optimistic performance (i.e., highest possible mF1 score thresholds per WSI) which is often reported in the literature.

|  | Validation mF1 Scores | | | Test mF1 Scores | | |
| --- | --- | --- | --- | --- | --- | --- |
| **Confidence Thresholds** | **Glomerulus** | **Arteriole** | **Artery** | **Glomerulus** | **Arteriole** | **Artery** |
| **0.3** | 0.723 | 0.471 | 0.590 | 0.758 | 0.416 | 0.691 |
| **0.5** | 0.772 | 0.512 | 0.629 | 0.802 | 0.436 | 0.772 |
| **0.7** | 0.819 | 0.502 | 0.657 | 0.838 | 0.418 | 0.734 |
| **0.9** | 0.873 | 0.349 | 0.655 | 0.860 | 0.255 | 0.675 |
| **Dynamic** | 0.896 | 0.490 | 0.704 | 0.860 | 0.363 | 0.663 |
| **Optimistic** | 0.911 | 0.577 | 0.769 | 0.895 | 0.493 | 0.789 |

**Table 2.** Final performance metrics after all WSI processing steps for the three object classes glomerulus, arteriole, artery. We note the high mean average precision scores for both glomerulus and artery. Although the arteriole class had lower performance, arteriole segmentation masks were deemed acceptable. Abbreviations: mIOU (mean intersection-over-union), mAP (mean average precision), mAR (mean average recall), mF1 (mean F1 score), mAS (mean average specificity). Validation and test performance were very similar.

|  | Validation Performance | | | | | Test Performance | | | | |
|---|---|---|---|---|---|---|---|---|---|---|
| Classes | mIOU | mAP | mAR | mF1 | mAS | mIOU | mAP | mAR | mF1 | mAS |
| **Glomerulus** | 0.818 | 0.880 | 0.919 | 0.896 | 0.993 | 0.759 | 0.900 | 0.829 | 0.860 | 0.995 |
| **Arteriole** | 0.342 | 0.531 | 0.488 | 0.490 | 0.996 | 0.230 | 0.537 | 0.311 | 0.363 | 0.996 |
| **Artery** | 0.565 | 0.739 | 0.679 | 0.704 | 0.995 | 0.510 | 0.757 | 0.640 | 0.663 | 0.989 |

As shown in Table 2, we achieved excellent performance for our segmentation module during both validation and testing testing with mean intersection-over-union (mIOU) of 0.818/0.759 for glomerulus, 0.342/0.230 for arterioles, 0.565/0.510 for artery; mean average precision (mAP) of 0.880/0.795 for glomerulus, 0.531/0.537 for arterioles, 0.739/0.757 for artery; mean average recall (sensitivity) of 0.919/0.829 for glomerulus, 0.488/0.311 for arteriole, 0.679/0.640 for artery; mean F1 (harmonic mean of precision and recall) of 0.896/0.860 for glomerulus, 0.490/0.363 for arteriole, 0.704/0.663 for artery; mean average specificity of 0.993/0.995 for glomerulus, 0.996/0.996 for arteriole, 0.995/0.989 for artery. Validation and testing performance were similar for all three compartment classes. Precision/recall curves are shown in Figure 2, exemplary predictions and mis-predictions in Figure 3.

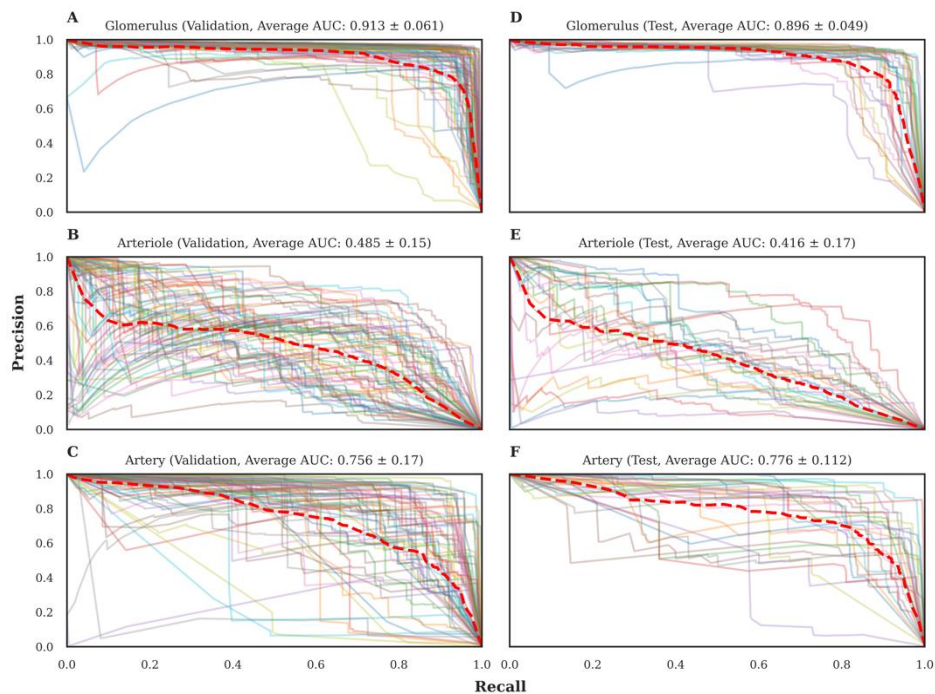

**Figure 2:** Comparison of class-wise performance between our validation set (**A** through **C**) and test set (**D** through **F**). Dotted red lines represent the average PR curve for the entire set, while each coloured line in the background represents an individual WSI. We noted a significant variability in performance for all classes, probably largely due to the unique characteristics of each input WSI.

# Discussion

Our segmentation model was trained on a dataset with great domain heterogeneity due to disease type and severity as well as staining diversity with up to 6 different stainings from multiple institutions, likely contributing to its robustness, even when validating and testing on previously unseen domains.

This accurate segmentation model should permit sufficiently accurate segmentation on large clinicopathological datasets from other nephropathology centres without or with minimal adjustment (e.g. by the MONAI Label active learning framework (https://monai.io/label.html, last accessed 1st of April 2023) plugged into QuPath [14]). Adding the object classes of cortex and medulla to our segmentation model should enable us to develop a full end-to-end pipeline of segmentation and compartment specific classification (with e.g., our MorphSet architecture [22]) for semi-supervised ML nephropathology classifiers with raw WSIs and their ground truth labels as input for binary binary, multi-class or regression tasks.

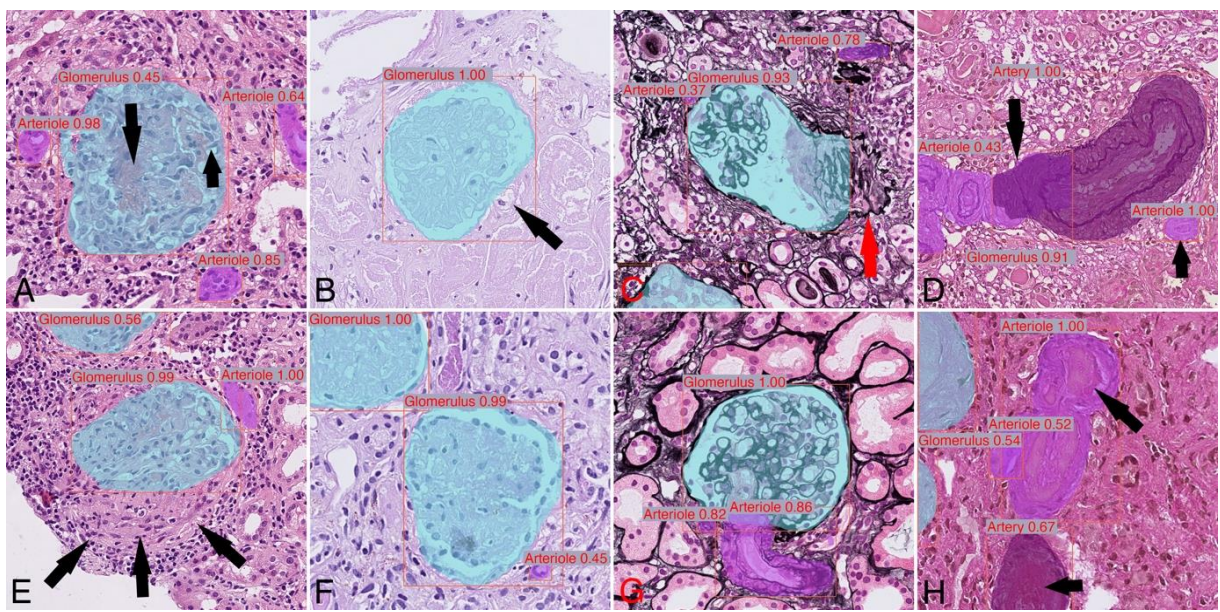

**Figure 3:** Exemplary prediction for the three object classes Artery (purple), Arteriole (pink), Glomerulus (turquoise). All bounding boxes also show the certainty for each instance prediction. Our instance segmentation module was trained and can be applied to all four main nephropathological stainings hematoxylin-eosin (HE) in A and E, periodic acid-Schiff (PAS) in B and F, Jones silver in C and G, trichrome in D and H (in this instance Elastica-van Gieson). A with perfect, very accurate predictions even on a severely distorted glomerular tuft with necrosis (long arrow) and cellular crescent (short arrow). B with an infarct-like necrosis of the glomerulus and the surrounding tissue in a biopsy with TMA; the infarcted glomerulus was predicted with certainty of 1.00. However, the feeding arteriole (arrow) was missed in this instance. In C a glomerulus with a crescent was predicted but for a small portion of the crescent (arrow). D with a transition (arrow) from artery to arteriole; the artery with hypoelastotic intimal fibrosis, typical for a thrombotic microangiopathy; note the tubule, mis-predicted with 1.00 certainty as an arteriole. E with a crescentic glomerular lesion (arrows) only partially predicted; note the accurately predicted glomerular tuft (turquoise). F with a partially infarcted but still accurately predicted glomerulus. G with the transition between predicted afferent arteriole, feeding into a glomerular tuft; note the missed efferent arteriole (arrow). H with an accurately predicted artery with a microthrombus (short arrow) and an arteriole branching-off from another arteriole, the latter also with a microthrombus (long arrow).

## Disclosure Statement

The authors have nothing to disclose.

## Acknowledgements

JUB is supported by the European Rare Kidney Disease Network (ERKNet), funded by the European Commission.

We acknowledge the excellent technical assistance of Bing He to Surya Seshan.

## References

1. Symmers, W.S.C., *Thrombotic microangiopathic haemolytic anaemia (thrombotic microangiopathy).* Br Med J 1952(II): p. 897-903.
2. Masias, C., S. Vasu, and S.R. Cataland, *None of the above: thrombotic microangiopathy beyond TTP and HUS.* Blood 2017. **129**(21): p. 2857-2863.
3. Furlan, M., R. Robles, M. Galbusera, G. Remuzzi, P.A. Kyrle, B. Brenner, M. Krause, I. Scharrer, V. Aumann, U. Mittler, M. Solenthaler, and B. Lammle, *von Willebrand factor-cleaving protease in thrombotic thrombocytopenic purpura and the hemolytic-uremic syndrome.* N Engl J Med 1998. **339**(22): p. 1578-84.
4. Levy, G.G., W.C. Nichols, E.C. Lian, T. Foroud, J.N. McClintick, B.M. McGee, A.Y. Yang, D.R. Siemieniak, K.R. Stark, R. Gruppo, R. Sarode, S.B. Shurin, V. Chandrasekaran, S.P. Stabler, H. Sabio, E.E. Bouhassira, J.D. Upshaw, Jr., D. Ginsburg, and H.M. Tsai, *Mutations in a member of the ADAMTS gene family cause thrombotic thrombocytopenic purpura.* Nature 2001. **413**(6855): p. 488-94.
5. Nester, C.M., T. Barbour, S.R. de Cordoba, M.A. Dragon-Durey, V. Fremeaux-Bacchi, T.H. Goodship, D. Kavanagh, M. Noris, M. Pickering, P. Sanchez-Corral, C. Skerka, P. Zipfel, and R.J. Smith, *Atypical aHUS: State of the art.* Mol Immunol 2015. **67**(1): p. 31-42.
6. Jokiranta, T.S., *HUS and atypical HUS.* Blood 2017. **129**(21): p. 2847-2856.
7. Laurence, J., H. Haller, P.M. Mannucci, M. Nangaku, M. Praga, and S. Rodriguez de Cordoba, *Atypical hemolytic uremic syndrome (aHUS): essential aspects of an accurate diagnosis.* Clin Adv Hematol Oncol 2016. **14 Suppl 11**(11): p. 2-15.
8. Campistol, J.M., M. Arias, G. Ariceta, M. Blasco, L. Espinosa, M. Espinosa, J.M. Grinyo, M. Macia, S. Mendizabal, M. Praga, E. Roman, R. Torra, F. Valdes, R. Vilalta, and S. Rodriguez de Cordoba, *An update for atypical haemolytic uraemic syndrome: Diagnosis and treatment. A consensus document.* Nefrologia 2015. **35**(5): p. 421-447.
9. Barisoni, L., K.J. Lafata, S.M. Hewitt, A. Madabhushi, and U.G.J. Balis, *Digital pathology and computational image analysis in nephropathology.* Nat Rev Nephrol 2020. **16**(11): p. 669-685.
10. Becker, J.U., D. Mayerich, M. Padmanabhan, J. Barratt, A. Ernst, P. Boor, P.A. Cicalese, C. Mohan, H.V. Nguyen, and B. Roysam, *Artificial intelligence and machine learning in nephropathology.* Kidney Int 2020. **98**(1): p. 65-75.
11. Lu, M.Y., D.F.K. Williamson, T.Y. Chen, R.J. Chen, M. Barbieri, and F. Mahmood, *Data-efficient and weakly supervised computational pathology on whole-slide images.* Nat Biomed Eng 2021. **5**(6): p. 555-570.
12. Zhang, H., Y. Meng, Y. Zhao, Y. Qiao, X. Yang, S.E. Coupland, and Y. Zheng. *DTFD-MIL: Double-Tier Feature Distillation Multiple Instance Learning for Histopathology Whole Slide Image Classification*. 2022; Available from: https://openaccess.thecvf.com/content/CVPR2022/papers/Zhang_DTFD-MIL_Double-Tier_Feature_Distillation_Multiple_Instance_Learning_for_Histopathology_Whole_CVPR_2022_paper.pdf.

13. Pfister, F., K. Amann, C. Daniel, M. Klewer, A. Buttner, and M. Buttner-Herold, *Characteristic morphological changes in anti-VEGF therapy-induced glomerular microangiopathy.* Histopathology 2018. **73**(6): p. 990-1001.
14. Bankhead, P., M.B. Loughrey, J.A. Fernandez, Y. Dombrowski, D.G. McArt, P.D. Dunne, S. McQuaid, R.T. Gray, L.J. Murray, H.G. Coleman, J.A. James, M. Salto-Tellez, and P.W. Hamilton, *QuPath: Open source software for digital pathology image analysis.* Sci Rep 2017. **7**(1): p. 16878.
15. Roufosse, C., N. Simmonds, M.C. Groningen, M. Haas, K.J. Henriksen, C. Horsfield, A. Loupy, M. Mengel, A. Perkowska-Ptasinska, M. Rabant, L.C. Racusen, K. Solez, and J.U. Becker, *A 2018 Reference Guide to the Banff Classification of Renal Allograft Pathology.* Transplantation 2018.
16. Otsu, N., *A Threshold Selection Method from Gray-Level Histograms.* EEE Transactions on Systems, Man, and Cybernetics 1979. **9**(1): p. 62-66.
17. Ronneberger, O., P. Fischer, and T. Brox. *U-Net: Convolutional Networks for Biomedical Image Segmentation*. 2015; Available from: https://arxiv.org/pdf/1505.04597.pdf.
18. He, K., G. Gkioxari, P. Dollár, and R. Girshick, *Mask R-CNN.* arXiv 2017.
19. Liu, Z., Y. Lin, Y. Cao, H. Hu, Y. Wei, Z. Zhang, S. Lin, and B. Guo. *Swin Transformer: Hierarchical Vision Transformer using Shifted Windows*. 2021; Available from: https://arxiv.org/pdf/2103.14030.pdf.
20. Loshchilov, I. and F. Hutter. *Decoupled Weight Decay Regularization*. 2019 [cited 2023; Available from: https://arxiv.org/pdf/1711.05101.pdf.
21. Cubuk, E.D., B. Zoph, M. D., V. Vasudevan, and Q.V. Le *AutoAugment: Learning Augmentation Strategies from Data*.
22. Cicalese, P.A., S.A. Rizvi, V. Wang, S. Patiibandla, P. Yuan, S. Zare, K. Moos, I. Batal, M.C. Clahsen-van Groningen, C. Roufosse, J. Becker, C. Mohan, and H.V. Nguyen, *MorphSet: Improving Renal Histopathology Case Assessment Through Learned Prognostic Vectors*, in *Medical Image Computing and Computer Assisted Intervention – MICCAI 2021 - 24th International Conference, Proceedings*. 2021: Virtual Online. p. 319-28.